\documentclass[]{spie}  

\usepackage{amsmath,amsfonts,amssymb}
\usepackage{graphicx}
\usepackage[colorlinks=true, allcolors=blue]{hyperref}
\usepackage{multirow}
\DeclareUnicodeCharacter{2212}{-}
\title{Automatic Tooth Segmentation from 3D Dental Model using Deep Learning: A Quantitative Analysis of what can be learnt from a Single 3D Dental Model}
\author[a,b]{Ananya Jana}
\author[a]{Hrebesh Molly Subhash}
\author[b]{Dimitris Metaxas}
\affil[a]{Colgate-Palmolive Company, Piscataway,USA}
\affil[b]{Dept. of Computer Science, Rutgers University, New Brunswick, USA}

\authorinfo{Further author information: (Send correspondence to Ananya Jana)\\Ananya Jana: E-mail: ananya.jana@rutgers.edu}

\begin{document} 
\maketitle

\begin{abstract}
3D tooth segmentation is an important task for digital orthodontics. Several Deep Learning methods have been proposed for automatic tooth segmentation from 3D dental models or intraoral scans. These methods require annotated 3D intraoral scans. Manually annotating 3D intraoral scans is a laborious task. One approach is to devise self-supervision methods to reduce the manual labeling effort. Compared to other types of point cloud data like scene point cloud or shape point cloud data, 3D tooth point cloud data has a very regular structure and a strong shape prior. We look at how much representative information can be learnt from a single 3D intraoral scan. We evaluate this quantitatively with the help of ten different methods of which six are generic point cloud segmentation methods whereas the other four are tooth segmentation specific methods. Surprisingly, we find that with a single 3D intraoral scan training, the Dice score can be as high as 0.86 whereas the full training set gives Dice score of 0.94. We conclude that the segmentation methods can learn a great deal of information from a single 3D tooth point cloud scan under suitable conditions e.g. data augmentation. We are the first to quantitatively evaluate and demonstrate the representation learning capability of Deep Learning methods from a single 3D intraoral scan. This can enable building self-supervision methods for tooth segmentation under extreme data limitation scenario by leveraging the available data to the fullest possible extent.\\
\end{abstract}
   \begin{figure*}[htbp]
    \centering
    \begin{minipage}{0.35\textwidth}
    \centering\includegraphics[width=\textwidth]{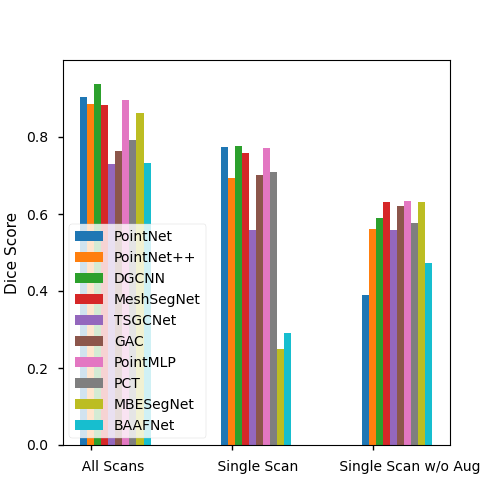} \\
    \textbf{Figure  1(a)}
    
    \end{minipage}
    \hspace{0.5cm}    
    \begin{minipage}{0.6\textwidth}
    \centering\includegraphics[width=\textwidth]{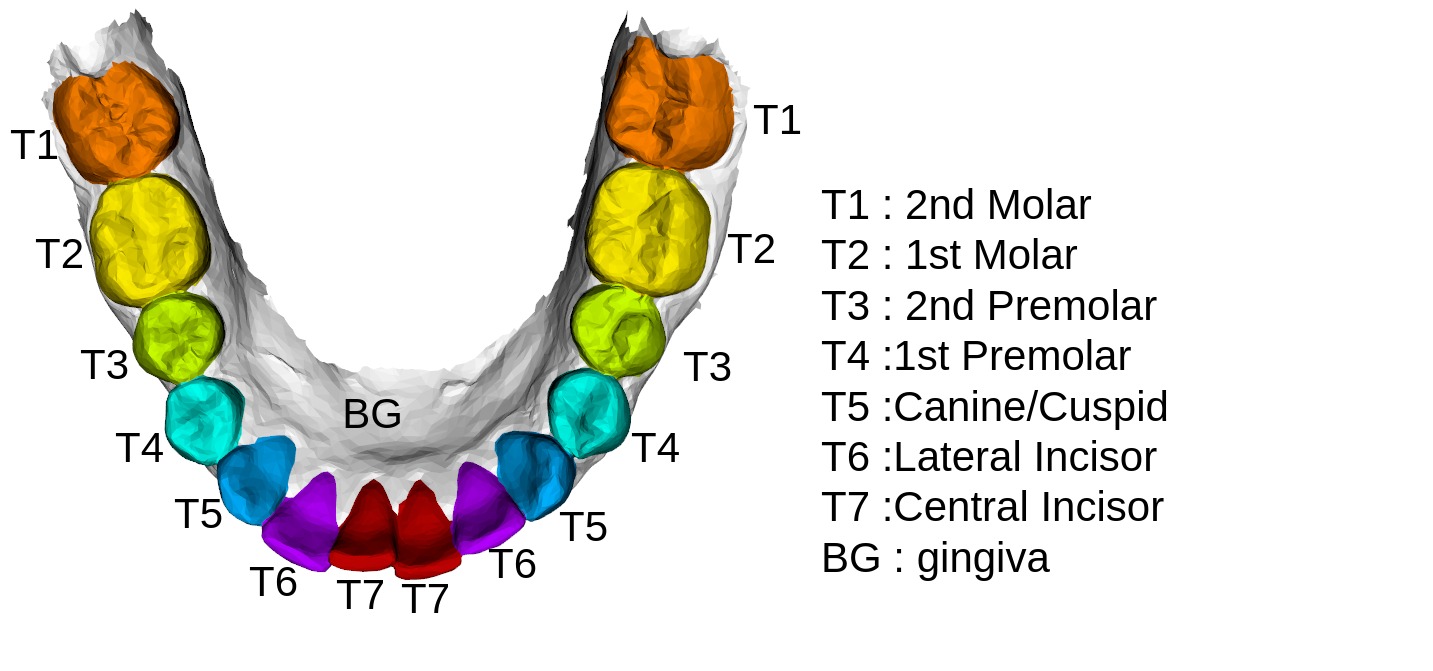} \\
    \textbf{Figure  1(b)}
    \end{minipage}
    \caption{Single-scan Representation Learning: We show that several Deep Learning methods can be
be trained(a) using a single 3D intraoral scan, such as
this Image (b)under suitable conditions.
}
    \label{fig:patVis}
\end{figure*}
\keywords{3D Tooth Segmentation, Tooth Point Cloud, Tooth Mesh, Intraoral scan segmentation, 3D Dental Model, Deep Learning, Machine Learning, Single Image Learning}

\section{INTRODUCTION}
\label{sec:intro} 
Intraoral scanners (IOS) are increasingly becoming an indispensable part of the digital dentistry due to its ability to quickly reconstruct the 3D surface. These cameras are widely being used in place of the traditional intraoral or dental cameras. While it may take hours to capture the pictures of each and every tooth of a subject using a dental camera, the intraoral scanners can get a scan in a matter of minutes. 
These 3D intraoral scans or 3D dental models can capture the 3D morphology, topology and also the color information in form of a point cloud or mesh. The 3D intraoral scans aid in the dental procedures like teeth aligner design, restoration design, smile beautification etc. Automatic 3D tooth point cloud segmentation is a crucial step for all these procedures.

Multiple Deep Learning based methods have been proposed for automatic tooth segmentation from the 3D intraoral scans in the recent years. We note that a majority of these methods\cite{lian2020deep, zhang2021tsgcnet, zhao20213d, li2022multi, zanjani2019deep, cui2021tsegnet, zheng2022teethgnn, liu2022heterogeneous, ma2020srf} are fully supervised with the exception of a few which are weakly supervised or semi-supervised\cite{qiu2022darch}. Lian et. al\cite{lian2020deep} proposed Meshsegnet which learns global information from the tooth point cloud using a pointnet similar network and the local information is learnt using different scale of adjacency matrices. Zhang et. al\cite{zhang2021tsgcnet} proposed a method which learns separately from the coordinates and the normals information by employing a parallel network. Qiu et. al\cite{qiu2022darch} proposed a method that leverages the fact that the teeth on the jaw have a symmetric structure and and imposes a shape prior on the curve connecting the tooth centroids. These methods are trained using annotated 3D intraoral scans (starting from 80\cite{zhang2021tsgcnet} to 4000\cite{qiu2022darch}). Manually annotating the 3D intraoral scans is a tedious process. Moreover, recently self-supervision methods have gained much attraction in the machine learning community. There has been great progress in self-supervision techniques. In self-supervision, pretext tasks are constructed automatically without needing any manual annotation on the data, which enables the network to learn useful features and utilize this knowledge in downstream tasks yielding impressive results even with limited data. In spite of the great benefits of self-supervision in natural images, and 3D point clouds like scene point cloud, shape point cloud etc, there is a lack of self-supervision methods in 3D tooth segmentation from intraoral scans. Self-supervision allows us to learn from limited data. and it is in this context that we investigate the representative power of a single intraoral scan. It is an important question to ask because unlike natural images which are more diverse, tooth scans have very strong shape and size prior - e.g. usually the jaw has the teeth aligned symmetrically and the teeth have relative size constraint e.g. a molar may not be smaller in volume in comparison to the incisor, there are two instances of teeth on each side of the jaw etc. Knowing how much information we can extract in supervised way from a single intraoral scan for 3D tooth segmentation can serve as a baseline for self-supervision i.e. networks will be able to learn a minimum of that much information via self-supervision.

We included ten different methods to evaluate the representative power of a single intraoral scan. We carefully pick the methods such that there are generic point cloud segmentation methods as well as tooth point cloud segmentation methods. The rationale behind this selection is to ensure that the analysis is generalizable across methods.
The reason we picked general segmentation methods is that although technically the tooth point cloud segmentation methods also operate on the point clouds, tooth point cloud is little different from the scene point clouds in the sense that most of the tooth point cloud segmentation methods actually operate on tooth mesh data but which is represented as point cloud data as we will see later in Section ~\ref{sec:processing}.
Our work is organized as the following - We first describe the dataset, its preprocessing, data augmentation method and then we introduce and outline the different generic and tooth specific point cloud segmentation methods. Next, we provide details of the experiments we performed and the results. Finally we conclude with a brief discussion of the results and their relevance.

\section{Method}
\subsection{Data Preprocessing}
\label{sec:processing}
We have a private de-identified dataset of 50 subjects. The dataset consists of 3D intraoral scans of the lower jaw of the subjects. The raw intraoral scan consists of more than 100000 meshes. We downsample the meshes to 16000 meshes using quadric downsampling method so that the topology of the raw intraoral scans are preserved. After downsampling each mesh cell is described with the help of the 24 Dimensions vector where the coordinates contribute to 12 dimensions and the normals contribute to the remaining 12 dimensions. The coordinates consist of the three vertices of the triangular mesh and the barycenter of the mesh whereas the normals correspond to the normals at each of the four coordinate points. The ground-truth annotations(autodesk MeshMixer) of all 3D dental models or intraoral scans is done following the clinical requirement and with the advice of professional dentists from our collaborative partner organization.An example of the annotated 3D intraoral scan is shown in Figure~\ref{fig:patVis}. Each scan has up to a maximum of 14 teeth (teeth count distribution is shown in The Table~\ref{tab:teethCount}) 

\begin{table*}[htbp]
\centering
\caption {The Tooth Count Distribution across Data Splits}
\begin{tabular}{|c|c|c|c|c | c | }
\hline
Data Split  & \#Teeth=9 & \#Teeth=11 & \#Teeth=12 &  \#Teeth=13 & \#Teeth=14\\ \hline
train set & 1 &  1 &  2 &  3 & 25 \\
val set & 0 & 0 & 2 & 1 & 5 \\
test set & 0 & 0 & 1 & 1 & 8\\
\hline
overall & 1 &  1 &  5 &  5 & 39 \\
\hline
\end{tabular}
\label{tab:teethCount}
\end{table*}  
    
\subsection{Data Augmentation}
To improve the models' generalization ability, the training and validation sets of the data split are augmented by combining 1) random rotation, 2) random translation, and
3) random rescaling (e.g., zooming in/out) of each 3D dental surface in reasonable ranges. Specifically, along each of the three axes in the 3D space, a training/validation surface has 50\% probability of translation with a displacement and zoom with a ratio uniformly sampled between [-10, 10] and [0.8, 1.2], respectively.Also, each training/validation surface has 50\% probability to be x-axis and y-axis and z-axis rotated with angles uniformly sampled between [−$\pi$, $\pi$]. The combination of these random operations simulated 40 “new” cases from each original surface. 

\subsection{Segmentation Methods}
\subsubsection{General Point Cloud Segmentation Methods}
In this section we take a look at the general point cloud segmentation methods. By the 'general' word we mean that the segmentation methods are independent of the type of the point cloud data e.g. scene point cloud, lidar point cloud or shape point cloud etc. \\ \\
\textbf{PointNet}\cite{qi2017pointnet} - It is a highly effective and efficient  deep learning based network architecture for point cloud processing tasks like object classification, part segmentation, to semantic parsing of scene. 
PointNet architecture grasps a good understanding of the global features of the point cloud.\\ \\
\textbf{PointNet++}\cite{qi2017pointnet++} - Qi et. al introduces a hierarchical feature learning alongwith the PointNet architecture and thus enabling a better understanding of the local neighborhood in point clouds.\\ \\
\textbf{DGCNN}\cite{wang2019dynamic} - This work introduces a network architecture for capturing the topological information from point clouds. Since point clouds are unordered set of points by nature/inherrently, they lack the topological information. DGCNN proposes a network module called EdgeConv which can be applied on dynamically computed graphs at different layers of the network. EdgeConv also captures the local neighborhood information.\\ \\
\textbf{PointMLP}\cite{ma2022rethinking} - Ma et. al shows that a pure residual network architecture for point clouds can achieve highly competitive performance without slowing down the network with expensive and sophisticated local geometric extractors/analyzers.\\ \\
\textbf{PCT}\cite{guo2021pct}  - Following the immense success of Transformer architectures in natural lanuage processing and natual image processing, Guo et. al proposes a transformer network for point cloud processing. It designs an input network carefully with farthest point sampling and nearest neighbor search such that network is benefited by both local and global structure of the point cloud.\\ \\ 
\textbf{BAAFNet}\cite{qiu2021semantic} - Qiu et. al proposes a network architecture which extracts the local context details by incorporating both geometric and semantic features in a bilateral fashion. They also process different resolutions of the pointcloud thereby reducing ambiguity and comprehensively interpreting the disntinctness of each point of the pointcloud.    

\subsubsection{Tooth Point Cloud Segmentation Methods}
In this section we take a look at the methods which are specifically designed for tooth segmentation. We will also illustrate how these methods are influenced by the general point cloud segmentation methods.\\ \\
\textbf{MeshSegNet}\cite{lian2020deep} - Lian et. al proposes a novel network architecture where graph-constrained learning modules help extracting multi-scale contextual features in a hierarchical fashion, and then densely assimilating the local-to-global geometric features for a comprehensive characterization of the tooth mesh cells for the segmentation task. This network borrows heavily from pointnet architecture but stands out by its use of adjacency matrix for understanding the local geometrics.\\ \\
\textbf{GAC}\cite{zhao20213d} - Zhao et. al proposes a network architecture which employs of two branches for extracting the fine-grained local information and the global information. The global feature branch is conceptually similar to PointNet architecture. The local information extractor branch introduces a module named LSAM . This module borrows concepts from EdgeConv from DGCNN architecture in that it learns edge weights of dynamically computed graphs but with the addition that the local context learning occurs through the use of both positional and semantic features. Graph Attention mechanism is used to summarize the information obtained from the dynamic graphs.\\ \\
\textbf{TSGCNet}\cite{zhang2021tsgcnet} - Zhang et. al proposes a two stream network where the two separate streams are dedicated for understanding the coordinates and the point normals. This method has elements in common with the GAC method in that it also makes use of graph attention layers for summarizing the graph weights information. But it makes the distinction that different types of information i.e. coordinates and normals should be initially processed separately to leverage them fully in characterizing the mesh cells.\\ \\
\textbf{MBESegNet}\cite{li2022multi} -  Li et. al proposes a network which is enabled to handle hierarchical and multi scale information. Similar to the concepts in BAAFNet\cite{qiu2021semantic}, this method also enhances the local feature bidirectionally with geometric and semantic information.\\ \\

\section{Experiments}
The task is tooth segmentation from the 3D dental model as C = 8 different semantic parts, indicating the central incisor (T7), lateral incisor (T6), canine/cuspid (T5), 1st premolar (T4), 2nd premolar (T3), 1st molar (T2), 2nd molar (T1), and background/gingiva (BG). All the experiments are conducted in a supervised fashion.
\subsection{Experiment Setting 1 (All Scans)}
We divide the dataset with a train-test-val split 32:8:10 for this experiment. The training set and validation set is augmented with the data augmentation method described previously. All the ten segmentation methods are trained on this entire dataset in fully supervised fashion. Our dataset consists of cases where there are missing teeth, mislaigned teeth, broken teeth etc.
\subsection{Experiment Setting 2 (Single Scan)}
In this setting we use only one data point or intraoral scan from the 32 subjects in training set of Experiment Setting 1 for training. The test and validation set remains the same as Experiment Setting 1. As in experiment setting 1, the trainset and validation set are augmented. the intraoral scan that has been selected has 14 teeth on it.
\subsection{Experiment Setting 3 (Single Scan w/o Augmentation)} This is similar to experiment setting 2 but with the exception that data augmentation has not been used. In this case the same sample is repeated multiple times to match the batch size of the training.
\subsection{Metric} We use four different metrics to measure the performance of the tooth segmentation methods. These metrics are Overall Accuracy(OA), Dice Score (DSC), Sensitivity (SEN) and Positive Predictive Value(PPV). For all the metrics, we take an average over all the classes.
\subsection{Training Details} The Experiments have been run for 400 epochs and the model yielding the best validation Dice score has been selected. All the models have been trained on RTX 8000 systems.

\begin{table*}[htbp]
\centering
\caption {The tooth segmentation results from then 10 different methods in terms of the Overall Accuracy and the Dice Score. The best results under each experimental setting is listed in bold.}
\begin{tabular}{|c|c|c|c|c | c |c| }
\hline
Method  & Exp &  OA & DSC & SEN &  PPV & Batch Time\\ \hline
\multirow{3}{*}{PointNet\cite{qi2017pointnet} } & All Scans & 0.9034 & 0.8873 & 0.8963 & 0.8971 & 3.325\\
& Single Scan & 0.7746 & 0.7161 & 0.7429 & 0.7193 & 57.295\\
& Single Scan w/o Aug & 0.39 & 0.1193 & 0.2075 & 0.8276 & 21.84\\
\hline
\multirow{3}{*}{PointNet++\cite{qi2017pointnet++}} & All Scans & 0.8868 & 0.8632 & 0.8695 & 0.8730 & 5.745\\ 
& Single Scan & 0.6923 & 0.5863 & 0.6205 & 0.6116 & 5.74\\
& Single Scan w/o Aug & 0.5620 & 0.4526 & 0.4819 & 0.4542 & 5.04\\
\hline
\multirow{3}{*}{DGCNN\cite{wang2019dynamic}} & All Scans & 0.9365 & 0.9206 & 0.9344 & 0.9244 & 2.998\\
& Single Scan &  0.7767 & 0.6985 & 0.7199 & 0.7120 & 2.66\\
& Single Scan w/o Aug & 0.5887 & 0.5012 & 0.5409 & 0.5089 & 25.30\\
\hline
\multirow{3}{*}{MeshSegNet\cite{lian2020deep}} & All Scans &  0.8824 & 0.8624 &  0.8902  & 0.8585 & 323.15\\
& Single Scan & 0.7586 & 0.7044 & 0.7394 & 0.7041 & 302.56\\
 & Single Scan w/o Aug &  0.6304 & 0.5360 & 0.5684 & 0.5425 & 4.124 \\
\hline
\multirow{3}{*}{MeshSegNet\cite{lian2020deep} with gco} & All Scans &  \textbf{0.9412} & 0.9288 &  0.9389  & 0.9217 & 421.30\\
& Single Scan & \textbf{0.8593} & 0.8173 & 0.8389 & 0.8121 & 353.23 \\
& Single Scan w/o Aug &  0.6307 &  0.7303 & 0.9530 & 0.7571 & 747.56\\
\hline
\multirow{3}{*}{TSGCNet\cite{zhang2021tsgcnet}} & All Scans & 0.7287 & 0.6541 & 0.6764 & 0.6688 & 3.97\\
& Single Scan & 0.5597 & 0.4413 & 0.4767 & 0.4612 & 6.65\\
& Single Scan w/o Aug  &  0.5581 &  0.4584 & 0.4959 & 0.4702 & 61.24\\
\hline
\multirow{3}{*}{GAC\cite{zhao20213d}} &  All Scans & 0.7634 & 0.7118  & 0.7207 & 0.7364 & 4.04\\
& Single Scan & 0.7002 & 0.6089 & 0.6399 & 0.6269 & 4.47  \\
& Single Scan w/o Aug  &  0.6215& 0.5323 & 0.5698 & 0.5430 & 3.28\\
\hline
\multirow{3}{*}{BAAFNet\cite{qiu2021semantic}} & All Scans &  0.7312 & 0.7125 & 0.7733 & 0.6922 & 35.87 \\
& Single Scan & 0.2919 &  0.2827 & 0.4146 & 0.2873  & 65.17 \\
& Single Scan w/o Aug & 0.4725 & 0.3461 & 0.3865 & 0.3585 & 86.02\\
\hline
\multirow{3}{*}{pointMLP\cite{ma2022rethinking}} & All Scans & 0.8972 & 0.8695 & 0.8789 & 0.8830 & 8.47 \\
& Single Scan & 0.7716 & 0.6848 & 0.7092 & 0.7021 & 10.13 \\
& Single Scan w/o Aug &  \textbf{0.6332} & 0.5180 & 0.5527 & 0.5190 & 53.00\\
\hline
\multirow{3}{*}{PCT\cite{guo2021pct} (coordinates only) } & All Scans & 0.7929 & 0.73941 & 0.7507 & 0.7668 & 4.51\\
& Single Scan & 0.7079 & 0.6396 & 0.6701 & 0.6536 & 21.83\\
& Single Scan w/o Aug & 0.5754 & 0.4962 & 0.5489 & 0.4790 & 1.94\\
\hline
\multirow{3}{*}{MBESegNet\cite{li2022multi}} & All Scans & 0.8614 & 0.8407 & 0.8436 & 0.8677 & 1.490\\
& Single Scan & 0.2488 & 0.1783 & 0.2135 & 0.2123 & 26.04\\
& Single Scan w/o Aug & 0.6304 & 0.5360 & 0.5683 & 0.5424 & 3.63\\
\hline
\end{tabular}
\label{tab:allres}
\end{table*}

\begin{figure*}[htbp]
\centering
\includegraphics[width=.8\textwidth]{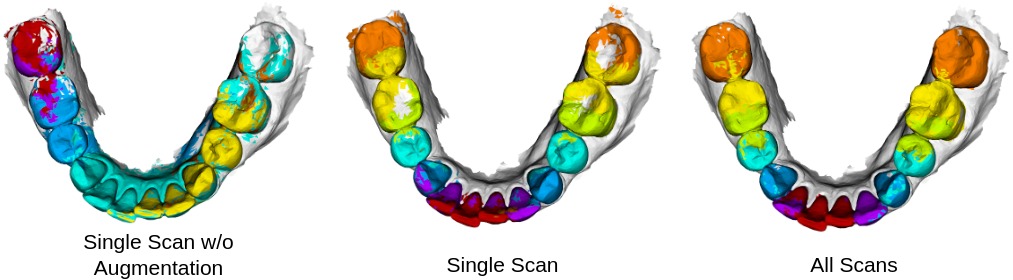}
\caption{Qualitative comparison of tooth labeling via DGCNN method when trained on all scans, single scan and single scan w/o augmentation. }
\label{fig:augVsNoAug}
\end{figure*}

\begin{table*}[htbp]
\centering
\caption {The segmentation results for the ten methods. All Scans, SS (Single Scan) and SSNA(Single Scan w/o Augmentation) denote the Experiment Setting 1, 2 and 3 respectively. }
\begin{tabular}{|c|c|c|c|c |c |c |c |c |c|}
\hline
Method  & Exp & BG & T1 & T2 & T3 &  T4 & T5 & T6 & T7  \\ \hline
\multirow{3}{*}{PointNet\cite{qi2017pointnet}} & All & 0.9496 & 0.7767 & 0.8248 & 0.8448 & 0.9207 & 0.9147 & 0.9184 & 0.9493\\
& SS & 0.8875 & 0.6826 & 0.6914 & 0.6906 & 0.7489 & 0.7503 & 0.6224 & 0.6554 \\
& SSNA & .0623 & .0100 & .0004 & .0007 & .0008 & .0009 &    .0011 & .0226 \\
\hline
\multirow{3}{*}{PointNet++\cite{qi2017pointnet++}} & All & 0.9409 & 0.8325 & 0.8407 & 0.8208 & 0.8756 & 0.8717 & 0.8420 & 0.8819 \\ 
& SS & 0.8462 & 0.6468 & 0.6873 & 0.6241 & 0.5648 & 0.3527 & 0.4283 & 0.5406 \\
& SSNA & 0.7700 & 0.5229 & 0.5055 & 0.3862 & 0.3937 & 0.3587 & 0.2911 & 0.3927 \\
\hline
\multirow{3}{*}{DGCNN\cite{wang2019dynamic}}  & All & 0.9772 & 0.8516 & 0.8642 & 0.9003 & 0.9614 & 0.9513 & 0.9191 & 0.9404 \\
& SS &  0.9298 & 0.5833 & 0.6542 & 0.6241 & 0.7261 & 0.7314 & 0.6669 & 0.6724 \\
& SSNA & 0.7855 & 0.4963 & 0.5589 & 0.3717 & 0.4401 & 0.4827 & 0.4031 & 0.4713 \\
\hline
\multirow{3}{*}{MeshSegNet\cite{lian2020deep}} & All  & 0.9353 & 0.7855 & 0.8065 & 0.8357 & 0.9154 & 0.8889 & 0.8667 & 0.8651\\
& SS & 0.8744 & 0.6034 & 0.6904 & 0.6866 & 0.7676 & 0.7014 & 0.6074 & 0.7043\\
& SSNA &  0.8002 & 0.5617 & 0.5754 & 0.4867 & 0.4775 & 0.4637 & 0.4114 & 0.5120\\
\hline
\multirow{3}{*}{" [gco] } & All  &0.9550 &  0.9470 & 0.9441 & 0.9167 & 0.9399 & 0.9152 & 0.9029 & 0.9103  \\
& SS & 0.9214 & 0.8825 & 0.8596 & 0.7984 & 0.7817 & 0.7710 & 0.7217 & 0.8026 \\
& SSNA & 0.0720 & 0.0014 & 1.0000 & 1.0000 & 1.0000 & 1.0000 & 1.0000 & 0.7694 \\
\hline
\multirow{3}{*}{BAAFNet\cite{qiu2021semantic}} & All & 0.7658 & 0.6824 & 0.7052 & 0.6797 & 0.7270 & 0.7221 & 0.6671 & 0.7515 \\
& SS & 0.0107 & 0.4805 & 0.4601 & 0.3010 & 0.3488 & 0.1947 & 0.1061 & 0.3598 \\
& SSNA &  0.7185 & 0.3357 & 0.4846 & 0.3987 & 0.2665 & 0.1777 & 0.0883 & 0.2995 \\
\hline 
\multirow{3}{*}{TSGCNet\cite{zhang2021tsgcnet}} & All  & 0.8366 & 0.6220 & 0.6626 & 0.6941 & 0.7060 & 0.6025 & 0.4926 & 0.6167 \\
& SS & 0.7411 & 0.5272 & 0.4681 & 0.3626 & 0.4146 & 0.2373 & 0.3595 & 0.4203\\
& SSNA & 0.7525 & 0.4532 & 0.5278 & 0.3458 & 0.4276 & 0.4235 & 0.2925 & 0.4447\\
\hline
\multirow{3}{*}{GAC\cite{zhao2006interactive}}  & All  & 0.8667 & 0.6680, & 0.6930 & 0.6526 & 0.7034 & 0.7201 & 0.6688 & 0.7220 \\
& SS & 0.8632 & 0.6677 & 0.6815 & 0.6181 & 0.5694 & 0.4655 & 0.4513 & 0.5547\\
& SSNA & 0.7941 & 0.5453 & 0.5882 & 0.5166 & 0.5494 & 0.4935 & 0.3076 & 0.4643\\
\hline
\multirow{3}{*}{pointMLP\cite{ma2022rethinking}}  & All &  0.9585 & 0.7876 & 0.8365 & 0.8532 & 0.8938 & 0.8773 & 0.8504 & 0.8987 \\
& SS & 0.9302 & 0.5905 & 0.6668 & 0.6556 & 0.7387 & 0.6731 & 0.5533 & 0.6708 \\
& SSNA &  0.8408 & 0.5530 & 0.5644 & 0.4970 & 0.5515 & 0.4037 & 0.3248 & 0.4090\\
\hline
\multirow{3}{*}{PCT\cite{guo2021pct} (coordinates only)}  & All  & 0.8773 & 0.6834 & 0.7477 & 0.7179 & 0.7593 & 0.7038 & 0.6865 & 0.7393 \\
& SS & 0.8304 & 0.6047 & 0.6516 & 0.5494 & 0.6070 & 0.5773 & 0.5719 & 0.7251\\
&  SSNA & 0.7325 & 0.5793 & 0.5401 & 0.4250 & 0.4936 & 0.4053 & 0.3520 & 0.4420\\
\hline
\multirow{3}{*}{MBESegNet\cite{li2022multi}}  & All &  0.9072 & 0.7050 & 0.7960 & 0.8111 & 0.8750 & 0.8942 & 0.8464 & 0.8910 \\
& SS & 0.2675 & 0.3806 & 0.2822 & 0.2205 & 0.1113 & 0.0780 & 0.0190 & 0.0678 \\
& SSNA & 0.8002 & 0.5617 & 0.5755 & 0.4866 & 0.4774 & 0.4636 & 0.4113 & 0.5119 \\
\hline
\end{tabular}
\label{tab:toothres}
\end{table*}

\begin{figure*}[htbp]
\centering
\includegraphics[width=1.0\textwidth]{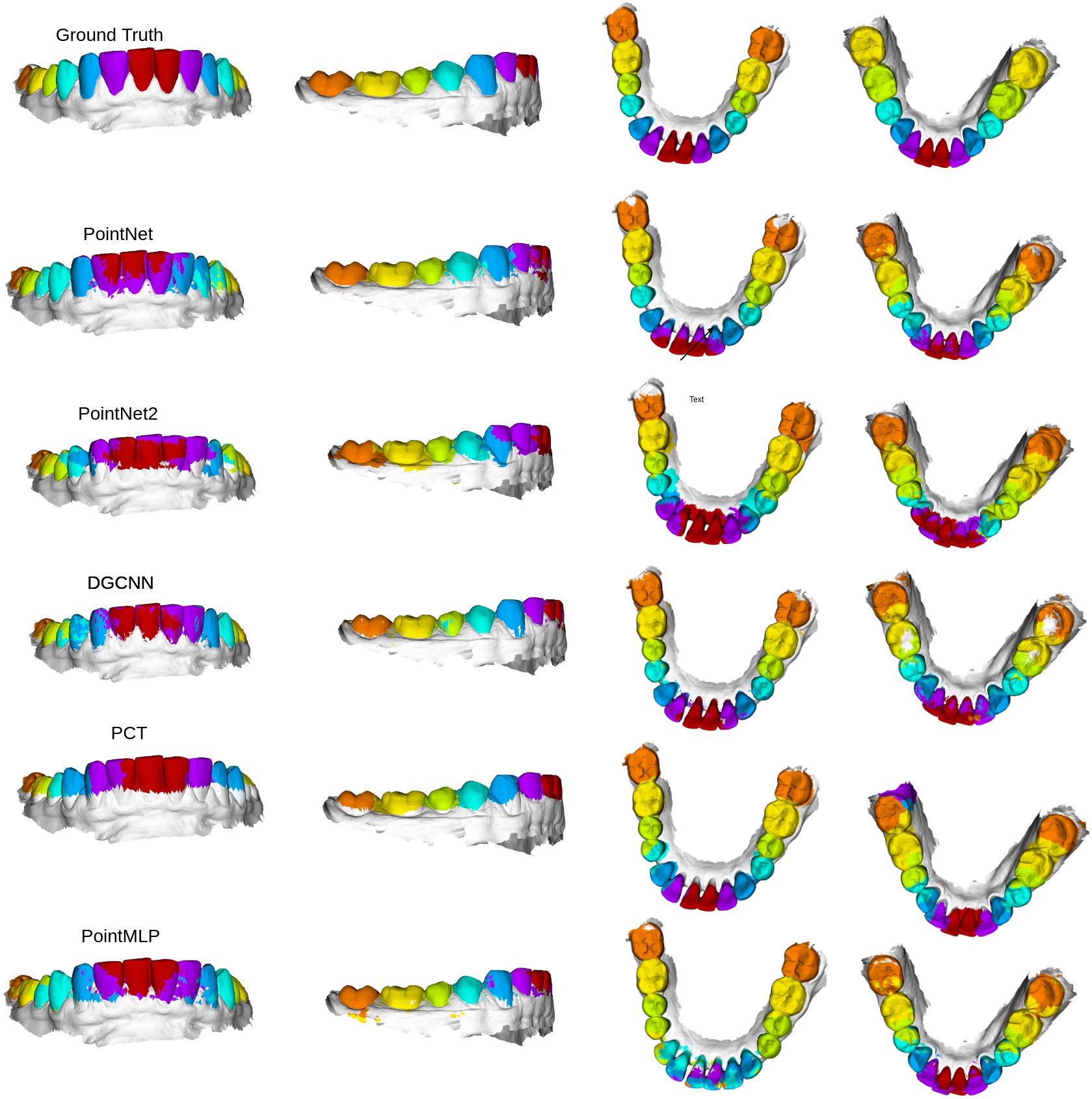}
\caption{The comparison of tooth labeling via the different methods trained on a single scan}
\label{fig:compareSingleTooth1}
\end{figure*}

\begin{figure*}[htbp]
\centering
\includegraphics[width=1.0\textwidth]{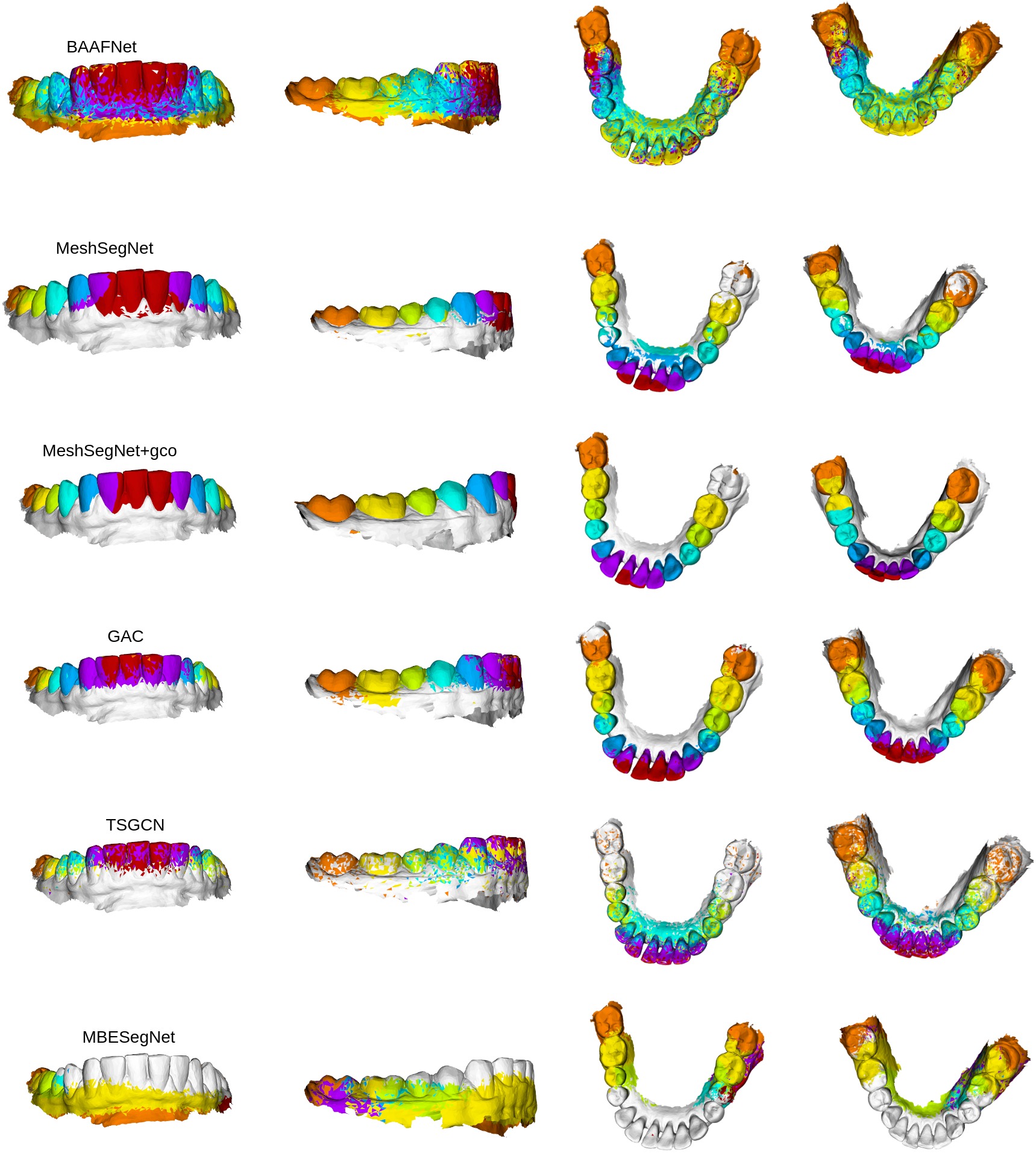}
\caption{The comparison of tooth labeling via the different methods trained on a single scan. }
\label{fig:compareSingleTooth2}
\end{figure*}

\section{Experimental Results}
\begin{figure*}[t]
\centering
\includegraphics[width=1.0\textwidth]{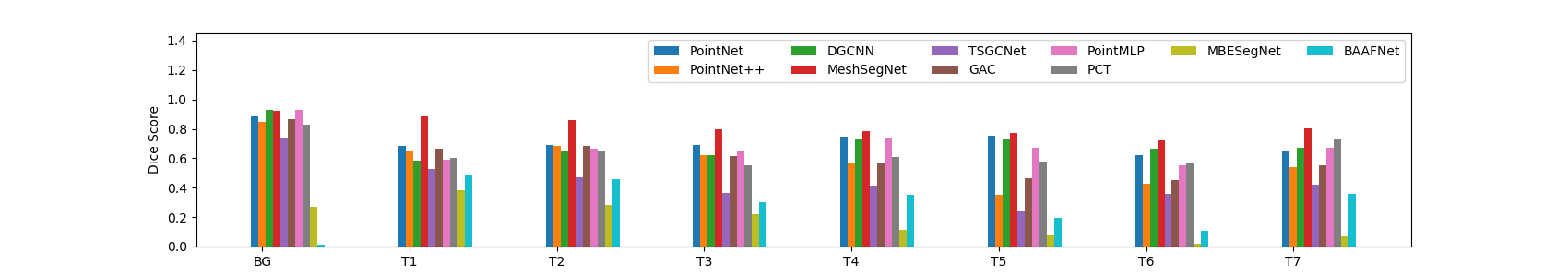}
\caption{The toothwise Dice score obtained via the different methods trained on a single scan}
\label{fig:toothDice}
\end{figure*}

The results of our experiments are listed in Table~\ref{tab:allres} and Table~\ref{tab:toothres}. The Table~\ref{tab:allres} shows the overall accuracy Dice score, sensitivity and positive predictive value of the different methods averaged across all the class labels under the three different experiment settings described previously i.e. All Scans, Single Scan and Single Scan w/o Augmentation. The Figure 1(a) shows the Dice score comparison of the different methods under the different settings. The Table~\ref{tab:toothres} and Figure~\ref{fig:toothDice} show the performance comparison of the ten different methods across the different class labels or in a toothwise fashion. 
The qualitative results of tooth labeling via the different methods trained on single scan is shown in Figure~\ref{fig:compareSingleTooth1} and Figure~\ref{fig:compareSingleTooth2}. For MeshSegNet\cite{lian2020deep} we included two different results - one with the graph cut post-processing(gco) and one without the post-processing. We observe that the graph cut post-processing which is a pluggable module can drive the Dice score higher but at the cost of additional computation power. We also note the effect of Data Augmentation in this experiment. All the methods except BAAFNet and MBESegNet have improvement in result with the data augmentation. This anomaly could probably attributed to the fact that the tooth meshes are being treated as tooth point clouds where the three vertices' coordinates are treated as features and predicting those from the tooth mesh triangle center might be challenging.  It can be noted that even without using any post-processing like graph cut(MeshSegNet), methods like DGCNN, PointNet, MeshSegNet and PointMLP achieve ~0.77 Dice score. Data augmentation seems to confuse the BAAFNet and MBESegNet networks. It is to be noted that at the time of submission of this work, implementations of TSGCNet, GAC and MBESegNet were not publicly available and hence we implemented those methods. Interestingly, PointNet performs the worst in terms of learning from a single scan when data augmentation is not used. It does not learn any information at all. The Figure~\ref{fig:augVsNoAug} shows a sample intraoral scan labelling via DGCNN method under the different experiment settings. The sample has incisor labels assigned to the molar teeth. This is possibly due to the lack of random rotation data augmentation, the DGCNN learning is based on the orientation of the input single scan on which it has been trained.  The PCT method is trained only on the coordinates and not the normals, yet succeeds in achieving good performance in a single scan setting. The single scan training for BAAFNet puts forward an interesting case where the mislabeled cells are not limited to a region. Rather they are scattered. This is  probably due to the fact that BAAFNet creates multiple resolutions from the original resolution of the tooth mesh, but the triangle vertices (and the normals) associated with the mesh cells do not change to accommodate the downsampling.

\section{Discussion}
Our work establishes quantitatively that it is possible to learn high amounts of information from a single 3D intraoral scan. This can be a starting point for devising self-supervision methods under extremely limited data annotation scenario. Moreover, the results also raise some interesting questions. We know that the Deep Learning based methods rely on large amounts of data, the general understanding is that the more data, the better results. But the current results compel us to think in the context of tooth segmentation: do we really need that much data? Or simply representative intraoral scans should be enough to learn from. The challenges of tooth segmentation from 3D intraorals scans, as commonly understood, are - misaligned teeth, missing teeth, broken teeth, crown etc. In that case, if we have representation of each challenge in form of an intraoral scan, can the model performance be driven to be still high? Another challenge in teeth segmentation is that not all subjects may have all the teeth e.g. kids may have 12 teeth on the gum. This can cause problems in the network by confusing T1 with T2. But if we look at an intraoral scan containing 14 teeth, in theory, it is possible to generate subsets e.g. 12 teeth intraoral scans from these scans using shape completion methods and then use the scan for training? Also, we can easily observe that tooth segmentation is in the intersection of part segmentation and instance segmentation problems. Naturally the question may arise, why not annotate half of the intraoral scan containing one instance of each teeth rather than the whole? Or we can translate this question to the more generic point clouds domain by asking, is it necessary to annotate an entire object if it is symmetric in nature? We conclude by saying that our work opens up many interesting avenues for future exploration by establishing the quantitative results of what can be learnt from a single intraoral scan.
\acknowledgments 
This work has been funded by the Colgate-Palmolive Company.  
\bibliography{report} 
\bibliographystyle{spiebib} 

\end{document}